\documentclass[runningheads]{llncs}
\usepackage[T1]{fontenc}
\usepackage{graphicx}
\usepackage{booktabs}

\usepackage{todonotes} % delete this probably
\usepackage{hyperref}
\usepackage{url}

\def\shortsetentity{\mathsf{E}}
\def\shortsetpartition{\mathsf{PA}}
\def\shortsetexhaustiveDecomposition{\mathsf{ED}}
\def\shortsetdisjointDecomposition{\mathsf{DD}}
\def\shortsetsubrelation{\mathsf{SR}}
\def\shortsetPredicate{\mathsf{PR}}

\makeatletter
\renewcommand\section{\@startsection{section}{1}{\z@}%
                       {-12\p@ \@plus -4\p@ \@minus -4\p@}%
                       {8\p@ \@plus 4\p@ \@minus 4\p@}%
                       {\normalfont\large\bfseries\boldmath
                        \rightskip=\z@ \@plus 8em\pretolerance=10000 }}
\makeatother

\begin{document}
\title{Translating SUMO-K to Higher-Order Set Theory}
%\thanks{Supported by organization x.}}
\author{Chad E. Brown\inst{1} \and
Adam Pease\inst{1,2}\orcidID{0000-0001-9772-1266} \and \\
Josef Urban\inst{1}\orcidID{0000-0002-1384-1613}}
\authorrunning{C. Brown, A. Pease, et al.}
\institute{
  Czech Technical University in Prague, Czech Republic\\
  \and
  Articulate Software, San Jose, CA, USA\\
 }
\maketitle              % typeset the header of the contribution
\begin{abstract}
  We describe a translation from a fragment of SUMO (SUMO-K) into
  higher-order set theory. The translation provides a formal
  semantics for portions of SUMO which are beyond first-order and
  which have previously only had an informal interpretation. It also
  for the first time embeds a large common-sense ontology into a very
  secure interactive theorem proving system.
  We further extend our
  previous work in finding contradictions in SUMO from first
  order constructs to include a portion of SUMO's higher order
  constructs. Finally, using the translation, we can create
  problems that can be proven using higher-order interactive and
  automated theorem provers. 
  This is tested in several systems and used to form a 
  corpus of higher-order common-sense reasoning problems.

   % \and Leo-III \and Eprover \and Vampire \and Zipperposition.}
\end{abstract}
\section{Introduction, Motivation and Related Work}

%\todo{C: This is just copied from the AITP submission. We should think about what's needed for this paper. AP- I'm happy with this as written.  It provides the appropriate context, it seems to me}
The Suggested Upper Merged Ontology (SUMO) \cite{np01,p11} is a comprehensive
ontology of around 20,000 concepts and 80,000 hand-authored logical statements
in a higher-order logic. It has an associated integrated development
environment called Sigma
\cite{ps14}\footnote{\url{https://www.ontologyportal.org}} that interfaces to
% leading
theorem provers such as E \cite{Sch02-AICOMM} and Vampire
\cite{Kovacs:2013:FTP:2958031.2958033}. In previous work on translating SUMO to
THF \cite{bp12}, a syntactic translation to THF was created but did not resolve
many aspects of the intended higher order semantics of SUMO.
% It is our objective in our current efforts

In this work, we 
% to
lay the groundwork for a new translation to
TH0, based on expressing SUMO in higher-order set theory. We believe
this will attach to SUMO a stronger set-theoretical interpretation that will
allow deciding more queries and provide better intuition for avoiding
contradictory formalizations. Once this is done, our plan is to train 
ENIGMA-style~\cite{JakubuvU17a,DBLP:conf/cade/ChvalovskyJ0U19,JakubuvU19,DBLP:conf/cade/JakubuvCOP0U20} query answering and
contradiction-finding~\cite{SchulzSutfliffeUrbanPease} AITP systems on such SUMO
problems and develop autoformalization~\cite{KaliszykUVG14,KaliszykUV15,DBLP:conf/itp/KaliszykUV17,DBLP:conf/mkm/WangKU18}
%DBLP:conf/cpp/WangBKU20}
methods
targeting common-sense reasoning based on SUMO. We believe that this is the most viable path towards common-sense reasoning that is both trainable, but also explainable and verifiable, providing an alternative to language models which come with no formal
guarantees.

In earlier work, we described \cite{ps14} how to translate SUMO to the strictly
first order language of TPTP-FOF \cite{psst10} and TF0 \cite{SutcliffeTFF,Pease2019,PeaseArxiv2023}. SUMO has an extensive type structure and all relations have type
restrictions on their arguments. Translation to TPTP FOF involved implementing a
sorted (typed) logic axiomatically in TPTP by altering all implications in SUMO
to contain type restrictions on any variables that appear.

%\todo{C: I added this to say something about the 35 challenge problems from 2010. AP - looks good to me}
In~\cite{PeaseSST10} 35 SUMO queries were converted into challenge problems for first-order automated
theorem provers.
In many cases, first-order ATPs can prove the corresponding problem.
However, some of the queries involve aspects of SUMO that go % require something
beyond
% a typical
first-order representation.
For example, one of the queries involves a term level binder ($\kappa$).
Several of the queries also involve \emph{row variables}, i.e., variables that should be instantiated
with a list of terms. We discuss here several such examples
% and related examples in order
to motivate the translation to higher-order set theory. We then embed SUMO into the Megalodon system,
providing, to our knowledge, the first representation of a large common-sense ontology within a very
  secure interactive theorem prover (ITP).
We then consider the higher-order problems one obtains via the translation.
This provides a set of challenge problems for higher-order theorem provers
that come from a very different source than formalized mathematics or
program verification.

The rest of the paper is organized as follows. In Section~\ref{sec:sumok} we introduce the SUMO-K fragment of SUMO,
an extension of the first-order fragment of SUMO.
Section~\ref{sec:translation} describes a translation from SUMO-K into
a higher-order set theory.
We have constructed interactive proofs of the translated form of
23 SUMO-K queries. We describe a few of these proofs in Section~\ref{sec:itp}.
From the interactive proofs we obtain 4880 ATP % premise selected
problems
and we describe the performance of higher-order automated theorem provers
on this problem set in Section~\ref{sec:atp}.
We describe plans to extend the work in Section~\ref{sec:futurework}
and conclude in Section~\ref{sec:concl}.
% At {\verb+http://grid01.ciirc.cvut.cz/~chad/sumo2set-0.9.tgz+}
Our code and problem set are available online.\footnote{\url{http://grid01.ciirc.cvut.cz/~chad/sumo2set-0.9.tgz}}

\section{The SUMO-K Fragment}\label{sec:sumok}

We define a fragment of SUMO we call SUMO-K.
Essentially, this extends the first-order fragment of SUMO with
support for row variables, variable arity functions and relations,
and the $\kappa$ class formation term binder.\footnote{SUMO classes should not be confused with set theoretic classes. Our use of ``class'' in this paper will always refer to SUMO classes.}
Elements of SUMO not included in SUMO-K are temporal, modal and probabilistic
operations.

We start by defining SUMO-K terms, spines (essentially lists of terms) and formulas.
%These extend the first-order fragment of SUMO with several features including $\kappa$-classes.
Formally, we have ordinary variables ($x$),
row variables ($\rho$) and constants ($c$).
We will also have signed rationals ($q$) represented by a decimal expression
with finitely many digits (i.e., those rationals expressible in such a way) as terms.
We mutually define the sets of SUMO-K terms $t$, SUMO-K spines $s$ and SUMO-K formulas $\psi$ as follows:
$$
\begin{array}{rcl}
  t & ::= & x | c | q | (x~s) | (c~s) | (\kappa x.\psi) | {\mathsf{Real}} | {\mathsf{Neg}} | {\mathsf{Nonneg}} | (t~+~t) | (t~-~t) | (t~*~t) | (t~/~t) \\
  s & ::= & t~s | \cdot | \rho | \rho~t\cdots~t \\
\psi & ::= & \bot | \top | (\lnot \psi) | (\psi \rightarrow \psi)
 | (\psi \land \psi)
 | (\psi \lor \psi)
 | (\psi \leftrightarrow \psi)
 | (\forall x.\psi)
 | (\exists x.\psi)
 | (\forall \rho.\psi)
 | (\exists \rho.\psi) \\
& | & (t = t) | ({\mathtt{instance}}~t~t)
 | ({\mathtt{subclass}}~t~t)
 | (t~\leq~t) | (t~<~t)
 | (c~s)
\end{array}
$$
The definition is mutually recursive since 
the term $\kappa x.\psi$
depends on the formula $\psi$.
Of course, $\kappa$, $\forall$ and $\exists$ are binders.
In practice, most occurrences of $\rho$ are at the end of the spine.
In some cases, however, extra arguments $t_1,\ldots, t_n$ occur after the $\rho$.
The idea is that $\rho$ will be list of arguments and $t_1,\ldots,t_n$ will be appended to
the end of that list. Note that at most one row variable can occur in a spine.

\subsection{Implicit Type Guards}

Properly parsing SUMO terms and formulas requires mechanisms
for inferring implicit type guards for variables
(interpreted conjunctively for $\kappa$ and $\exists$
and via implication for $\forall$).
Free variables in SUMO assertions are implicitly universally quantified
and are restricted by inferred type guards, as described in~\cite{ps14}.
% For simplicity, we assume all type guards and implicit quantifiers
% have already been inferred (as in~\cite{ps14}) before beginning the translation.
In previous translations targeting first-order logic,
relation and function variables are instantiated during the translation
(treating the general statement quantifying over relations and functions
as a macro to be expanded).
Since the current translation will leave these as variables,
we must also deal with type guards that are not known until
the relation or function is instantiated.

\subsection{Variable Arity Relations and Functions}

Consider the SUMO relation ${\mathsf{partition}}$, declared as follows:
\begin{verbatim}
(instance partition Predicate)
(instance partition VariableArityRelation)
(domain partition 1 Class)
(domain partition 2 Class)
\end{verbatim}
The last three items indicate that ${\mathsf{partition}}$ has variable arity
with at least 2 arguments, both of which are intended to be classes.
If there are more than 2 arguments, the remaining arguments are also intended to be classes.
In general, the extra optional arguments of a variable arity relation or function
are intended to have the same domain as the last required argument.
We will translate ${\mathsf{partition}}$ to a set that encodes not only
when the relation should hold, but also its domain information, its minimum arity
and whether or not it is variable arity.

Two other variable arity relations (with the same arity and type information as ${\mathsf{partition}}$)
are ${\mathsf{exhaustiveDecomposition}}$ and ${\mathsf{disjointDecomposition}}$.
%For the sake of brevity, we use $P$ for ${\mathsf{partition}}$, $E$ for
%${\mathsf{exhaustiveDecomposition}}$ and $D$ for ${\mathsf{disjointDecomposition}}$.
The following is an example of a SUMO-K assertion relating these concepts:
%$$\forall\rho.P~\rho\rightarrow E~\rho~\land~D~\rho$$
$$\forall\rho.{\mathsf{partition}}~\rho\rightarrow {\mathsf{exhaustiveDecomposition}}~\rho~\land~{\mathsf{disjointDecomposition}}~\rho.$$
Previous translations to first-order expanded this assertion into several facts
for different possible arities (using different predicates ${\mathsf{partition}}_3$,
${\mathsf{partition}}_4$, etc.), up to some limit.
The following is an example of a partition occurring in Merge.kif\footnote{Merge.kif is the main SUMO ontology file. While Merge.kif evolves over time, we work with a fixed version of the file from January 2023. Latest versions of it and all the other files that make up SUMO are available at \url{https://github.com/ontologyportal/sumo}}
with 6 arguments:
{\footnotesize{
\begin{verbatim}
(partition Word Noun Verb Adjective Adverb ParticleWord)
\end{verbatim}
}}
From this one should be able to infer the following query:
\begin{example}[${\mathsf{wordex}}$]\label{ex:wordex}
{\footnotesize{
\begin{verbatim}
(query (exhaustiveDecomposition
          Word Noun Verb Adjective Adverb ParticleWord))
\end{verbatim}
}}
\end{example}
However, the corresponding first-order problem will not be provable unless
the limit on the generated arity is at least 6.
Our translation into set theory will free us from the need to know such limits
in advance.

\subsection{Quantification over Relations}

Merge.kif includes assertions that quantify over relations.
The following is an example of such an assertion:
{\footnotesize{
\begin{verbatim}
(=> (and (subrelation ?REL1 ?REL2) (instance ?REL1 Predicate)
         (instance ?REL2 Predicate) (?REL1 @ROW))
    (?REL2 @ROW))
\end{verbatim}
}}

In previous first-order translations such assertions are instantiated
with all $R$ and $R'$ where $({\mathsf{subrelation}}~R~R')$ is asserted.
One of the 35 problems from~\cite{PeaseSST10} (${\mathsf{TQG22}}$) makes use of
the SUMO assertion that ${\mathsf{son}}$ is a subrelation of ${\mathsf{parent}}$
and the macro expansion style of first-order translation is sufficient to
handle this example.
However, the macro expansion approach is insufficient to handle
hypothetical subrelation assertions.
The following is an example of a query creating a hypothetical subrelation assertion:
\begin{example}[${\mathsf{TQG22alt4}}$]\label{ex:tqg22alt4}
{\footnotesize{
\begin{verbatim}
(query (=> (exists (?X) (employs ?X ?X))
           (not (subrelation employs uses))))
\end{verbatim}
}}
\end{example}
During the process of answering this query we will assume ${\mathsf{employs}}$
is a subrelation of ${\mathsf{uses}}$
and then must instantiate the general assertion about subrelations
with ${\mathsf{employs}}$ and ${\mathsf{uses}}$.
Our translation to set theory will permit this.

\subsection{Kappa Binders}\label{ref:kappa}

One of the 35 queries from~\cite{PeaseSST10} (${\mathsf{TQG27}}$) has the following local assumption
making use of a $\kappa$-binder.
\begin{example}\label{ex:tqg27}
  The example ${\mathsf{TQG27}}$ includes three assertions: (A1)
  ${\mathsf{instance}}~{\mathsf{Planet}}~{\mathsf{Class}}$, (A2) 
  ${\mathsf{subclass}}~{\mathsf{Planet}}~{\mathsf{AstronomicalBody}}$,
  and (the one with a $\kappa$-binder)\\ (A3)
  ${\mathsf{instance}}~o~(\kappa p.{\mathsf{instance}}~p~{\mathsf{Planet}}~\land~{\mathsf{attribute}}~p~{\mathsf{Earthlike}}).$\\
  The query is (Q) ${\mathsf{instance}}~o~{\mathsf{Planet}}$.
%{\footnotesize{
%(instance Planet27-1 Class)
%(subclass Planet27-1 AstronomicalBody)
%(instance Object27-1 (KappaFn ?PLANET (and (instance ?PLANET Planet27-1) (attribute ?PLANET Earthlike))))
%(query (instance Object27-1 Planet27-1))
%}}
\end{example}
%The assumption with the $\kappa$-binder has the form
%$${\mathsf{instance}}~o~(\kappa p.{\mathsf{instance}}~p~{\mathsf{Planet}}~\land~{\mathsf{attribute}}~p~{\mathsf{Earthlike}}).$$
%The query is simply ${\mathsf{instance}}~o~{\mathsf{Planet}}$
The query should easily follow by eliminating the $\kappa$-abstraction.
The first-order problem generated in~\cite{PeaseSST10} drops the assumption with the $\kappa$-abstraction (A3),
making the problem unlikely to be provable (at least not for the intended reason).
Our translation to set theory will handle $\kappa$-binders and the translation
of this problem will be provable in the set theory.

\subsection{Real Arithmetic}\label{sec:real}

Six of the 35 examples from~\cite{PeaseSST10} involve some real arithmetic.
Two simple example queries are the following:
\begin{example}[${\mathsf{TQG3}}$]\label{ex:tqg3}
{\footnotesize{
\begin{verbatim}
(instance Number3-1 NonnegativeRealNumber)
(query (not (instance Number3-1 NegativeRealNumber)))
\end{verbatim}
}}
\end{example}
\begin{example}[${\mathsf{TQG11}}$]\label{ex:tqg11}
{\footnotesize{
\begin{verbatim}
(query (equal 12 (MultiplicationFn 3 4)))
\end{verbatim}
}}
\end{example}
For the sake of brevity we represent the first problem
as having one local constant $n$, one local assumption
${\mathsf{instance}}~n~{\mathsf{Nonneg}}$
and the query (conjecture) $\lnot ({\mathsf{instance}}~n~{\mathsf{Neg}})$.
%The second query is represented as $12 = 3~*~4$.
We will translate signed rationals with a finite decimal expansion
to real numbers represented as sets.\footnote{We use a fixed construction of the reals, but the details of this are not relevant here.}
We will also translate ${\mathsf{Real}}$ to be equal to the set of reals $\Re$.
and translate the operations $+$, $-$, $*$, $/$, $<$ and $\leq$ to have
the appropriate meaning when applied to two reals.\footnote{For simplicity, our set theoretic division is a total function returning $0$ when the denominator is $0$.}
We then translate ${\mathsf{Neg}}$ to $\{x\in \Re|x < 0\}$
and ${\mathsf{Nonneg}}$ to $\{x\in \Re|0 \leq x\}$.
Using the properties of the set theoretic encoding, the translated queries above are
set theoretic theorems.

In addition to direct uses of arithmetic as in the examples above,
arithmetic is also often used to check type guard information.
This is due to the fact that a spine like $t_1\,t_2\,\rho$
will use subtraction to determine that under some constraints
the $i^{th}$ element of the corresponding list will be the $(i-2)^{nd}$ element
of the list interpreting $\rho$.

\section{Translation of SUMO-K to Set Theory}\label{sec:translation}

Our translation maps terms $t$ to sets.
The particular set theory we use is
higher-order Tarski-Grothendieck as described in~\cite{BrownPak19}.\footnote{Tarski-Grothendieck is a set theory in which there are universes modeling ZFC set theory. These set theoretic universes should not be confused with the universe of discourse ${\mathsf{Univ1}}$ introduced below.}
% We will often present the images of translated SUMO items
% using Megalodon syntax.
% Megalodon is an interactive theorem prover for higher-order set theory
% and is the successor to the Egal system also described in~\cite{BrownPak19}.
The details of this set theory are not important here.
We only note that we have $\in$, $\subseteq$
(which will be used to interpret SUMO's ${\mathtt{instance}}$
and ${\mathtt{subclass}}$)
and that we have the ability to $\lambda$-abstract variables
to form terms at higher types.
The main types of interest are $\iota$ (the base type of sets),
$o$ (the type of propositions),
$\iota\to\iota$ (the type of functions from sets to sets)
and
$\iota\to o$ (the type of predicates over sets).
When we say SUMO terms $t$ are translated to sets, we mean they
are translated to terms of type $\iota$ in the higher-order set theory.

Spines $s$ are essentially lists of sets (of varying length).
We translate them to functions of type $\iota\to\iota$
but only use them when restricted to arguments $n\in \omega$.
We also maintain the invariant that the function returns the empty set
on all but finitely many $n\in\omega$.
A function ${\mathsf{listset}}:(\iota\to\iota)\to\iota$ gives
a set theoretic representation of the list by restricting its domain to $\omega$.
To avoid confusion with the empty set
being on the list, we tag the elements of the list to ensure they are nonempty
(and then untag them when using them).
Let ${\mathsf{I}}:\iota\to\iota$ be such a tagging function (injective on the universe of sets)
and ${\mathsf{U}}:\iota\to\iota$ be an untagging function.
We define ${\mathsf{nil}}:\iota\to\iota$
to be constantly $\emptyset$
and ${\mathsf{cons}}:\iota\to (\iota\to\iota)\to\iota\to\iota$
to take $x$ and $l$ to the function mapping $0$ to ${\mathsf{I}}~x$
and $i+1$ to $l~i$ for $i\in\omega$.
We define a function ${\mathsf{len}}:(\iota\to\iota)\to\iota$
by $\lambda l.\{i\in\omega|l~i\not=\emptyset\}$
giving us the length of the list (assuming it is a list).
Informally, a spine like $t_0\cdots t_{n-1}$ is a function
taking $i$ to $I(t'_i)$ for each $i\in\{0,\ldots,n-1\}$
where $t'_i$ is the set theoretic value of $t_i$
and $I$ is the tagging operation.
%Each SUMO spine $s$ will be translated to a term of type $\iota$ following this idea.

The translation of a SUMO formula $\psi$ can be thought of either as a set
(which should be one of the sets $0$ or $1$) or as a proposition.
We also sometimes coerce between type $\iota$ and $o$
by considering the sets $0$ and $1$ to be sets corresponding to false and true.
Let ${\mathsf{P}}:\iota\to o$ be $\lambda X.\emptyset\in X$
and let ${\mathsf{B}}:o\to\iota$ be $\lambda p.{\mathsf{if}}~p~then~{\mathsf{1}}~else~{\mathsf{0}}$.
We use these functions as coercions between $\iota$ and $o$ when necessary.

Before describing the translation in more detail, we give a few more simple examples
to explain various aspects of the translation and motivate our choices.

Let ${\mathsf{Univ1}}$ be a set, intended to be a universe
of discourse in which most (but not all) targets of interpretation
for $t$ will live.
Specifically we will map the SUMO-type ${\mathsf{Class}}$ to the
set $\wp~{\mathsf{Univ1}}$ (the power set of the universe).
For all SUMO-types except the four special cases
${\mathsf{Class}}$, ${\mathsf{SetOrClass}}$, ${\mathsf{Abstract}}$
and ${\mathsf{Entity}}$
to be sets in $\wp~{\mathsf{Univ1}}$.
Consequently, if a SUMO object is an instance of some class
other than ${\mathsf{Class}}$, ${\mathsf{SetOrClass}}$, ${\mathsf{Abstract}}$
and ${\mathsf{Entity}}$,
we will know that the object is a member of ${\mathsf{Univ1}}$.
Due to this we choose to translate $\kappa$-binders
using simple separation bounded by ${\mathsf{Univ1}}$.
Reconsidering ${\mathsf{TQG27}}$ discussed in Subsection~\ref{ref:kappa}
we translate
${\mathsf{instance}}~o~(\kappa p.{\mathsf{instance}}~p~{\mathsf{Planet}}~\land~{\mathsf{attribute}}~p~{\mathsf{Earthlike}})$
to a set theoretic proposition of the form
$o\in\{p\in{\mathsf{Univ1}} | \cdots p\in {\mathsf{PLANET}}~\land~\cdots\}$
(only partially specified at the moment).
From this set theoretic proposition we can easily derive $o\in {\mathsf{PLANET}}$
to solve the set theoretic version of ${\mathsf{TQG27}}$.

As mentioned above, ${\mathsf{partition}}$ is a variable arity relation
of at least arity $2$ where every argument must be of SUMO-type ${\mathsf{Class}}$.
We will translate ${\mathsf{partition}}$ to a set
${\shortsetpartition}$ containing multiple pieces of information.
The behavior of ${\shortsetpartition}$ as a relation is captured
by the results one obtains by applying it to a set encoding a list of sets
(via a set theoretic operation ${\mathsf{ap}}:\iota\to\iota\to\iota$).
We can apply an abstract function ${\mathsf{arity}}:\iota\to\iota$
to obtain the minimum arity of ${\shortsetpartition}$.
We can apply an abstract predicate ${\mathsf{vararity}}:\iota\to o$
to encode that ${\shortsetpartition}$ has variable arity.
Likewise we can apply an abstract ${\mathsf{domseq}}:\iota\to\iota\to\iota$
to ${\shortsetpartition}$ and an $i\in \omega$ to recover the intended
domain of argument $i$ of ${\shortsetpartition}$.
These extra pieces of information are important to determine type guards
in the presence of function and relation arguments.

In the specific case of ${\mathsf{partition}}$ the translation yields
a set ${\shortsetpartition}$ such that
${\mathsf{arity}}~{\shortsetpartition} = 2$,
${\mathsf{vararity}}~{\shortsetpartition}$ is true
and for $i\in\{0,1,2\}$,
${\mathsf{domseq}}~{\shortsetpartition}~i = \wp~{\mathsf{Univ1}}$.
The value of ${\mathsf{domseq}}~{\shortsetpartition}~2$ determines
the intended domain of all remaining (optional) arguments of the relation.
(Note that SUMO indexes the first argument by $1$ while in the set theory the first argument is indexed by $0$.)
The SUMO assertion
{\footnotesize{
\begin{verbatim}
(partition Word Noun Verb Adjective Adverb ParticleWord)
\end{verbatim}
}}
\noindent
translates to the set theoretic statement
$$
\begin{array}{c}
  {\mathsf{P}}~({\mathsf{ap}}~{\shortsetpartition}~({\mathsf{listset}}~({\mathsf{cons}}~{\mathsf{Word}}~({\mathsf{cons}}~{\mathsf{Noun}}~({\mathsf{cons}}~{\mathsf{Verb}}~({\mathsf{cons}}~{\mathsf{Adjective}}\\
  ({\mathsf{cons}}~{\mathsf{Adverb}}~({\mathsf{cons}}~{\mathsf{ParticleWord}}~{\mathsf{nil}})))))))).
\end{array}
$$

Recall the SUMO-K assertion
$$\forall\rho.{\mathsf{partition}}~\rho\rightarrow {\mathsf{exhaustiveDecomposition}}~\rho~\land~{\mathsf{disjointDecomposition}}~\rho.$$
In this case the translation also generates type guards for the row variable $\rho$.
Let ${\shortsetpartition}$, ${\shortsetexhaustiveDecomposition}$ and
${\shortsetdisjointDecomposition}$ be the sets
corresponding to the SUMO constants
${\mathsf{partition}}$, ${\mathsf{exhaustiveDecomposition}}$ and ${\mathsf{disjointDecomposition}}$.
Essentially, the assertion should only apply to $\rho$ when $\rho$ has at least length 2
and every entry is a (tagged) class.
The translated set theoretic statement (with type guards) is
$$
\begin{array}{c}
  \forall \rho:\iota\to\iota.{\mathsf{dom\_of}}~({\mathsf{vararity}}~{\shortsetpartition})~({\mathsf{arity}}~{\shortsetpartition})~({\mathsf{domseq}}~{\shortsetpartition})~\rho\\
  \rightarrow{\mathsf{dom\_of}}~({\mathsf{vararity}}~{\shortsetexhaustiveDecomposition})~({\mathsf{arity}}~{\shortsetexhaustiveDecomposition})~({\mathsf{domseq}}~{\shortsetexhaustiveDecomposition})~\rho\\
  \rightarrow{\mathsf{dom\_of}}~({\mathsf{vararity}}~{\shortsetdisjointDecomposition})~({\mathsf{arity}}~{\shortsetdisjointDecomposition})~({\mathsf{domseq}}~{\shortsetdisjointDecomposition})~\rho\\
  \rightarrow {\mathsf{P}}~({\mathsf{ap}}~{\shortsetpartition}~\rho)
  \rightarrow {\mathsf{P}}~({\mathsf{ap}}~{\shortsetexhaustiveDecomposition}~\rho)
  \land {\mathsf{P}}~({\mathsf{ap}}~{\shortsetdisjointDecomposition}~\rho)
\end{array}
$$
The statement above makes use of a new definition: ${\mathsf{dom\_of}}:o\to\iota\to(\iota\to \iota)\to(\iota\to\iota)\to o$.
The first argument of ${\mathsf{dom\_of}}$ is a proposition encoding whether or not the function
or relation is variable arity. In this case, all three of the propositions are variable arity
(with the same typing information for all three).
In the variable arity case ${\mathsf{dom\_of}}~\top~n~D~\rho$
is defined to be ${\mathsf{dom\_of\_varar}}~n~D~\rho$
where
${\mathsf{dom\_of\_varar}}:\iota\to(\iota\to \iota)\to(\iota\to\iota)\to o$,
$n$ is the minimum arity, $D$ is the list of domain information
and $\rho$ is the list we are requiring to satisfy the guard.
${\mathsf{dom\_of\_varar}}~n~D~\rho$ is defined to hold
if the following three conditions hold:
\begin{enumerate}
\item $n\subseteq {\mathsf{len}}~\rho$ ($\rho$ has at least length $n$)
\item $\forall i\in n, {\mathsf{U}}~(\rho~i) \in D~i$ and
\item $\forall i \in {\mathsf{len}}~\rho, n \subseteq i \rightarrow {\mathsf{U}}~(\rho~i)\in D~n$.
\end{enumerate}
For fixed arity, ${\mathsf{dom\_of}}$ is defined via a simpler
${\mathsf{dom\_of\_fixedar}}$ condition.

Another SUMO assertion about partitions is
{\footnotesize{
\begin{verbatim}
(=> (partition ?SUPER ?SUB1 ?SUB2) (partition ?SUPER ?SUB2 ?SUB1))
\end{verbatim}
}}
In this case there are three ordinary (nonrow) variables needing type
guards in the translation.
Roughly speaking, ${\mathsf{domseq}}~{\shortsetpartition}$
has the information we need, but in general we must modify it to
be appropriate for variable arity relations.
For this reason ${\mathsf{domseqm}}: \iota\to\iota\to\iota$
is defined to be
$$\lambda r i.{\mathsf{if}}~{\mathsf{vararity}}~r~{\mathsf{then}}~{\mathsf{domseq}}~r~({\mathsf{if}}~i \in~{\mathsf{arity}}~r~{\mathsf{then}}~i~{\mathsf{else}}~{\mathsf{arity}}~r)~{\mathsf{else}}~{\mathsf{domseq}}~r~i.$$
The translated statement is
$$
\begin{array}{c}
\forall X Y Z. X\in {\mathsf{domseqm}}~{\shortsetpartition}~0
\rightarrow Y \in {\mathsf{domseqm}}~{\shortsetpartition}~1
\rightarrow Z \in {\mathsf{domseqm}}~{\shortsetpartition}~2\\
\rightarrow Z \in {\mathsf{domseqm}}~{\shortsetpartition}~1
\rightarrow Y \in {\mathsf{domseqm}}~{\shortsetpartition}~2\\
\rightarrow {\mathsf{P}} ({\mathsf{ap}}~{\shortsetpartition}~({\mathsf{cons}}~X~({\mathsf{cons}}~Y~({\mathsf{cons}}~Z~{\mathsf{nil}})))))\\
\rightarrow {\mathsf{P}} ({\mathsf{ap}}~{\shortsetpartition}~({\mathsf{cons}}~X~({\mathsf{cons}}~Z~({\mathsf{cons}}~Y~{\mathsf{nil}}))))).
\end{array}
$$

A simpler translation for handling type guards in this example
could avoid the use of ${\mathsf{dom\_of}}$ and ${\mathsf{domseqm}}$ and instead
look up the arity and typing information for ${\mathsf{partition}}$, etc.
This translation would not work in general since SUMO assertions
quantify over relations, in which case the particular type guards
are not known until the relation variables are instantiated.
Consider the SUMO-K formula
$$
\begin{array}{c}
  \forall R_1 R_2.\forall \rho.
          {\mathsf{subrelation}}~R_1~R_2~\land~{\mathsf{instance}}~R_1~{\mathsf{Predicate}} \rightarrow
          {\mathsf{instance}}~R_2~{\mathsf{Predicate}} \\
          \rightarrow R_1~\rho\rightarrow R_2~\rho.
\end{array}
$$
This translates to the set theoretic proposition
$$
\begin{array}{c}
  \forall R_1 R_2:\iota.\forall \rho:\iota\to\iota.
  R_1 \in {\mathsf{domseqm}}~{\shortsetsubrelation}~0
  \rightarrow
  R_2 \in {\mathsf{domseqm}}~{\shortsetsubrelation}~1
  \rightarrow\\
  R_1 \in {\shortsetentity}
  \rightarrow
  R_2 \in {\shortsetentity}
  \rightarrow
      {\mathsf{dom\_of}}~({\mathsf{vararity}}~R_1)~({\mathsf{arity}}~R_1)~\rho\\
  \rightarrow
      {\mathsf{dom\_of}}~({\mathsf{vararity}}~R_2)~({\mathsf{arity}}~R_2)~\rho\\
  \rightarrow
      {\mathsf{P}}~({\mathsf{ap}}~{\shortsetsubrelation}~({\mathsf{cons}}~R_1~({\mathsf{cons}}~R_2~{\mathsf{nil}})))
      \land
      R_1 \in {\shortsetPredicate}
      \land
      R_2 \in {\shortsetPredicate}
      \land
          {\mathsf{P}}~({\mathsf{ap}}~R_1~\rho)\\
          \rightarrow {\mathsf{P}}~({\mathsf{ap}}~R_2~\rho)
\end{array}
$$
where ${\shortsetentity}$, ${\shortsetsubrelation}$ and ${\shortsetPredicate}$
are the sets corresponding to the SUMO constants
${\mathsf{Entity}}$,
${\mathsf{subrelation}}$
and ${\mathsf{Predicate}}$.
Here the type guards on $\rho$ depend on $R_1$ and $R_2$.
Two special cases are the type guards $R_i\in {\shortsetentity}$
which are derived from the use of $R_i$ as the first argument
of ${\mathsf{instance}}$.

\subsection{The Translation}\label{subsec:translation}

We now describe the translation itself.
A first pass through the SUMO files given
records the typing information from ${\mathsf{domain}}$,
${\mathsf{range}}$, ${\mathsf{domainsubclass}}$,
${\mathsf{rangesubclass}}$
and ${\mathsf{subrelation}}$
assertions.
A finite number of secondary passes determines which
names will have variable arity (either due to a direct
assertion or due to being inferred to be in a variable arity class).\footnote{In practice with the current Merge.kif file, a single secondary pass suffices, but in general one might need an extra pass to climb the class hierarchy.}

The final pass translates the assertions, and this is our focus here.
Each SUMO-K assertion is a SUMO-K formula $\varphi$ which may have free variables in it.
Thus if we translate the SUMO-K formula $\varphi$ into the set theoretic proposition
$\varphi'$, then the translated assertion will be
$$\forall x_1\cdots x_n.G_1\to\cdots G_m\to\varphi'$$
where $x_1,\ldots,x_n$ are the free variables in $\varphi$
and $G_1,\ldots,G_m$ are the type guards for these free variables.
Note that some of these free variables may be for spine variables (i.e., row variables)
and may have type $\iota\to \iota$. Such variables may also have type guards.

SUMO-K variables $x$ translate to themselves where after translation
$x$ is a variable of type $\iota$ (ranging over sets).
For SUMO-K constants $c$ we choose a name $c'$ and declare this as having
type $\iota$.
Rational numbers $q$ with a finite decimal expansion
are translated to the set calculating the quotient of the base ten
numerator divided by the appropriate power of $10$.
For example, 11.2 would be translated to the term $1 * 10^2 + 1 * 10 + 2$
divided by $10$
(where $1$, $2$ and $10$ are the usual finite ordinals
and exponentiation by finite ordinals is defined by recursion).
When a variable or constant is applied to a spine we translate the spines
and use ${\mathsf{ap}}$.
As mentioned in Section~\ref{sec:real} ${\mathsf{Real}}$ is translated to the set $\Re$,
${\mathsf{Neg}}$ is translated to $\{x\in \Re|x < 0\}$
and ${\mathsf{Nonneg}}$ is translated to $\{x\in \Re|0 \leq x\}$.
The other arithmetical constructs are translated to sets,
but we assume special properties such as
$$\forall x y\in\Re.{\mathsf{ap}}~{\mathsf{ADD}}~({\mathsf{cons}}~x~({\mathsf{cons}}~y~{\mathsf{nil}})) = x + y,$$
$$\forall x y\in\Re.{\mathsf{ap}}~{\mathsf{MULT}}~({\mathsf{cons}}~x~({\mathsf{cons}}~y~{\mathsf{nil}})) = x \cdot y$$
and
$$\forall x y\in\Re.{\mathsf{P}}~({\mathsf{ap}}~({\mathsf{LESSTHAN}}~({\mathsf{cons}}~x~({\mathsf{cons}}~y~{\mathsf{nil}})))) = (x < y).$$
\begin{itemize}
\item $(x~s)$ translates to $({\mathsf{ap}}~x~({\mathsf{listset}}~s'))$
  where $s'$ is the result of translating the SUMO-K spine $s$.
\item $(c~s)$ translates to $({\mathsf{ap}}~c'~({\mathsf{listset}}~s'))$
  where $s'$ is the result of translating the SUMO-K spine $s$
  and $c'$ is the chosen set as a counterpart to the SUMO-K constant $c$.
  Arithmetical operations are handled the same way.
\end{itemize}
The only remaining case for terms is $\kappa$ binder terms.
\begin{itemize}
\item We translate $(\kappa x.\psi)$ to
  $$\{x\in{\mathsf{Univ1}}~|~ G_1\land \ldots G_m\land \psi'\}$$
  where $G_1,\ldots,G_m$ are generated type guards for $x$
  and $\psi'$ is the result of translating the SUMO-K formula $\psi$
  to a set theoretic proposition.
  Note that $x$ ranges over ${\mathsf{Univ1}}$.
\end{itemize}

The translations of spines is relatively straightforward.
\begin{itemize}
\item The SUMO-K spine $(t~s)$ is translated to
  the list one gets by applying ${\mathsf{cons}}$ to $I~t'$ onto $s'$
  where $t'$ is the translation of $t$ and $s'$ is the translation of $s$.
\item A spine variable $\rho$ is translated to itself (a variable of type $\iota\to\iota$).
\item In the case $\rho~t_1~\ldots~t_n$ we translate $\rho$ to itself (a variable of type $\iota\to\iota$)
  and translate each $t_i$ to a set $t'_i$ and return
  the function that returns $\rho~j$ given $j < {\mathsf{len}}~\rho$
  and returns $t'_i$ given ${\mathsf{len}}~\rho~+~i$
  (appending the two lists).
\item The empty spine is translated to ${\mathsf{nil}}$.
\end{itemize}

We consider each case of a SUMO-K formula.
The usual logical operators are translated as the corresponding operators:
\begin{itemize}
\item $\bot$ and $\top$ translate simply to $\bot$ and $\top$.
\item $(\lnot~\psi)$ translates to $\lnot \psi'$
  where $\psi$ is a SUMO-K formula which translates to the set theoretic proposition $\psi'$.
\item $(\psi~\rightarrow~\xi)$ translates to $\psi' \to \xi'$
  where $\psi$ and $\xi$ are SUMO-K formulas translate to the set theoretic propositions $\psi'$ and $\xi'$.
\item $(\psi~\leftrightarrow~\xi)$ translates to $\psi' \leftrightarrow \xi'$
  where $\psi$ and $\xi$ are SUMO-K formulas translate to the set theoretic propositions $\psi'$ and $\xi'$.
\item Theoretically, $\psi\land \xi$ translates to $\psi'\land \xi'$.
  Practically speaking in SUMO-K conjunction is $n$-ary
  so it is more accurate to state that
  $({\mathsf{and}}~\psi_1~\cdots \psi_n)$ translates to $\psi_1' \land\cdots\land \psi'_n$
  where $\psi_1,\ldots,\psi_n$ are SUMO-K formulas translate to the set theoretic propositions
  $\psi_1',\ldots,\psi_n'$.
\item Again, theoretically $\psi\lor \xi$ translates to $\psi'\lor \xi'$.
  Practically,
  $({\mathsf{or}}~\psi_1~\cdots \psi_n)$ translates to $\psi_1' \lor\cdots\lor \psi'_n$
  where $\psi_1,\ldots,\psi_n$ are SUMO-K formulas translate to the set theoretic propositions
  $\psi_1',\ldots,\psi_n'$.
\item Theoretically, $\forall x.\psi$ translates to $\forall x.G_1 \to\cdots\to G_m\to \psi'$
  where $\psi'$ is the result of translating $\psi$ and $G_1,\ldots,G_m$ are the generated type guards for $x$.
  Practically speaking, SUMO-K allows several variables to be universally quantified at once,
  so it is more accurate to say
  $({\mathsf{forall}}~(x_1\ldots x_n)~\psi)$ translates to
  $\forall x_1\ldots x_n.G_1 \to\cdots\to G_m\to \psi'$
  where $x_1,\ldots,x_n$ are variables,
  $G_1,\ldots,G_m$ are the generated type guards for these variables
  and $\psi'$ is the set theoretic proposition obtained by translating $\psi$.
\item $\forall \rho.\psi$ is translated similarly, but with type guards for the row variable $\rho$.
\item Again, theoretically $\exists x.\psi$ translates to $\exists x.G_1 \land\cdots\land G_m\land \psi'$,
  where $\psi'$ is the set theoretic proposition obtained by translating the SUMO-K formula $\psi$,
  but generalized to handle quantifying multiple variables.
\item $\exists \rho.\psi$ is translated similarly, but with type guards for the row variable $\rho$.
\item $(t_1=t_2)$ translates to $t_1' = t_2'$
  where $t_1$ and $t_2$ are SUMO terms which translate to sets $t_1'$ and $t_2'$.
\end{itemize}
We use set membership and inclusion to interpret ${\mathtt{instance}}$ and ${\mathtt{subclass}}$.
\begin{itemize}
\item $({\mathtt{instance}}~t_1~t_2)$ translates to $t_1'\in t_2'$
  where $t_1$ and $t_2$ are SUMO terms which translate to sets $t_1'$ and $t_2'$.
\item $({\mathtt{subclass}}~t_1~t_2)$ translates to $t_1'\subseteq t_2'$
  where $t_1$ and $t_2$ are SUMO terms which translate to sets $t_1'$ and $t_2'$.
\end{itemize}

\section{Interactive Proofs of Translated SUMO Queries}\label{sec:itp}

The motivating set of examples were the 35 example queries from~\cite{PeaseSST10}, now 
expanded\footnote{\url{https://github.com/ontologyportal/sumo/tree/master/tests}}.
Six of the original examples involve temporal reasoning. We omit these for the
moment, leaving a future translation to handle temporal and modal reasoning.
9 questions involve too many arguments for the existing first order
translation with macro expansion to work, but which are handled by our new translation.
One problem requires negation by failure. Among the remaining problems,
5 require some arithmetical reasoning, which use preexisting translations to standard first-order logic (FOF)
and to an extension of first-order logic with arithmetic (TFF).
For the remaining problems, the results of (at least) 5 were still not provable
by the ATPs Vampire or E within a 600 second timeout.

We carefully looked at the set theoretic translation of 13 of the problems
that were too difficult for first-order provers (for any of the above reasons
other than the use of temporal or modal reasoning).
We either did an interactive proof
or found slight modifications of the problem that could be interactively proven.
The interactive proofs were done in Megalodon (the
successor to the Egal system~\cite{BrownPak19}).
One advantage of having such a translation is the ability to attempt
interactive proofs and recognize what may be missing from Merge.kif or
the original query.
We also did interactive proofs of 4 problems that the first order provers could prove.
We additionally included the 6 problems dealing with variable arity
and row variables (e.g., Example~\ref{ex:wordex}).
In total we have 23 SUMO-K queries translated to set theoretic statements
that have been interactively proven.
We briefly describe some of the interactive proofs here.

An example with a particularly simple proof is ${\mathsf{TQG27}}$ (Example~\ref{ex:tqg27}), the
example with a $\kappa$-binder.
The assertion with the $\kappa$-binder translates to
the set theoretic proposition
$$
\begin{array}{rl}
  o\in \{p\in {\mathsf{Univ1}}| & p\in {\shortsetentity} \land p\in {\mathsf{domseqm}}~{\mathsf{attribute}}~0\land p\in {\mathsf{Planet}}\\
  & \land {\mathsf{P}}~({\mathsf{ap}}~{\mathsf{attribute}}~({\mathsf{listset}}~({\mathsf{cons}}~p~({\mathsf{cons}}~{\mathsf{Earthlike}}~{\mathsf{nil}}))))\}.
\end{array}
$$
The query translates simply to $o\in{\mathsf{Planet}}$.

When interactively proving the translated query in Megalodon, we
are free to use statements coming from three sources:
set theoretic propositions already previously proven in Megalodon
(or are axioms of Tarski-Grothendieck),
propositions resulting from the translation of formulas in Merge.kif,
and propositions resulting from translating formulas local to the example.
In this case we only need two propositions: the translated formula
local to the example given above and
one known set theoretic proposition of the form:
$$\forall X:\iota.\forall P:\iota\to o.\forall x:\iota.x\in \{x\in X|P~x\}\to x\in X\land P~x.$$
From the two propositions we easily obtain the conjunction
$$
\begin{array}{c}
  o\in {\mathsf{Univ1}} \land o\in {\shortsetentity} \land o\in {\mathsf{domseqm}}~{\mathsf{attribute}}~0\land o\in {\mathsf{Planet}}\\
  \land {\mathsf{P}}~({\mathsf{ap}}~{\mathsf{attribute}}~({\mathsf{listset}}~({\mathsf{cons}}~o~({\mathsf{cons}}~{\mathsf{Earthlike}}~{\mathsf{nil}})))).
\end{array}
$$
After this first step, a series of steps eliminate the conjunctions
until we have the desired conjunct $o\in {\mathsf{Planet}}$.

Another relatively simple example is ${\mathsf{TQG11}}$ (Example~\ref{ex:tqg11})
in which we must essentially prove $12$ is $3\cdot 4$.
To be more precise we must prove
$$1\cdot 10 + 2 = {\mathsf{ap}}~{\mathsf{MULT}}~({\mathsf{listset}}~({\mathsf{cons}}~3~({\mathsf{cons}}~4~{\mathsf{nil}}))).$$
As mentioned in Section~\ref{subsec:translation} the translation adds the
proposition
$$\forall x y\in\Re.{\mathsf{ap}}~{\mathsf{MULT}}~({\mathsf{cons}}~x~({\mathsf{cons}}~y~{\mathsf{nil}})) = x \cdot y$$
which will be useful here.
In the interactive proof, we first prove a claim that every natural number (finite ordinal)
is a real number (i.e., $\omega\subseteq\Re$, which is true for the representation of the reals being used).
This claim is then used to prove $3\in \Re$ and $4\in \Re$.
This allows us to reduce the main goal to proving
$1\cdot 10 + 2 = 3 \cdot 4$.
This goal is then proven by an unsurprising sequence of
rewrites using equations defining the behavior of $+$ and $\cdot$ on finite
ordinals. (Many details are elided here, such as the fact that there are
actually two different operations $+$, one on reals and one only on finite ordinals
and that they provably agree on finite ordinals.)

We next consider the proof of the translation of Example~\ref{ex:tqg22alt4}.
The set theoretic proposition resulting from translating the query is
$$
\begin{array}{c}
  (\exists x. x\in {\mathsf{domseqm}}~{\mathsf{employs}}~0 \land x\in {\mathsf{domseqm}}~{\mathsf{employs}}~1\\
  \land {\mathsf{P}}~({\mathsf{ap}}~{\mathsf{employs}}~({\mathsf{listset}}~({\mathsf{cons}}~x~({\mathsf{cons}}~x~{\mathsf{nil}}))))) \\
  \to \lnot {\mathsf{P}}~({\mathsf{SR}}~({\mathsf{listset}}~({\mathsf{cons}}~{\mathsf{employs}}~({\mathsf{cons}}~{\mathsf{uses}}~{\mathsf{nil}})))).
\end{array}
$$
We begin the interactive proof by proving the following sequence of claims:
\begin{enumerate}
\item ${\mathsf{listlen}}~{\mathsf{nil}} = 0$.
\item $\forall X.\forall R:\iota\to\iota.\forall n.{\mathsf{nat\_p}}~n\to {\mathsf{listlen}}~R=n\to {\mathsf{listlen}}~({\mathsf{cons}}~X~R) = {\mathsf{ordsucc}}~n$.
\item $\forall y.\lnot{\mathsf{vararity}}~y\to\forall i.{\mathsf{domseqm}}~y~i = {\mathsf{domseq}}~y~i$.
\item $\forall y.\lnot{\mathsf{vararity}}~y\to\forall x i. x \in {\mathsf{domseq}}~y~i \to x \in {\mathsf{domseqm}}~y~i$.
\item $\forall X.\forall R:\iota\to\iota.{\mathsf{cons}}~X~R~0 = {\mathsf{I}}~X$
\item $\forall n.{\mathsf{nat\_p}}~n\to\forall X.\forall R:\iota\to\iota.{\mathsf{cons}}~X~R~({\mathsf{ordsucc}}~n) = {\mathsf{R}}~n$.
\end{enumerate}
We can then rewrite ${\mathsf{domseqm}}~{\mathsf{employs}}$
into ${\mathsf{domseq}}~{\mathsf{employs}}$.
Starting the main body of the proof,
we assume we have an $x$
such that
$x\in {\mathsf{domseq}}~{\mathsf{employs}}~0$,
$x\in {\mathsf{domseq}}~{\mathsf{employs}}~1$
and
${\mathsf{P}}~({\mathsf{ap}}~{\mathsf{employs}}~({\mathsf{listset}}~({\mathsf{cons}}~x~({\mathsf{cons}}~x~{\mathsf{nil}}))))$.
We further assume
${\mathsf{P}}~({\mathsf{SR}}~({\mathsf{listset}}~({\mathsf{cons}}~{\mathsf{employs}}~({\mathsf{cons}}~{\mathsf{uses}}~{\mathsf{nil}})))$
and prove a contradiction.
Using the translated Merge.kif type information from ${\mathsf{employs}}$
we can infer $x$ is an autonomous agent and an object.
Likewise we can infer ${\mathsf{employs}}$ is a predicate and a relation,
and the same for ${\mathsf{uses}}$.
The contradiction follows from two claims:
${\mathsf{P}}~({\mathsf{ap}}~{\mathsf{uses}}~({\mathsf{cons}}~x~({\mathsf{cons}}~x~{\mathsf{nil}})))$
and
$\lnot {\mathsf{P}}~({\mathsf{ap}}~{\mathsf{uses}}~({\mathsf{cons}}~x~({\mathsf{cons}}~x~{\mathsf{nil}})))$.

We first prove ${\mathsf{P}}~({\mathsf{ap}}~{\mathsf{uses}}~({\mathsf{cons}}~x~({\mathsf{cons}}~x~{\mathsf{nil}})))$.
We locally let ${\mathsf{ROW}}$ be ${\mathsf{cons}}~x~({\mathsf{cons}}~x~{\mathsf{nil}})$
and use the claims above prove from ${\mathsf{ROW}}~0 = {\mathsf{I}}~x$,
${\mathsf{ROW}}~1 = {\mathsf{I}}~x$,
${\mathsf{U}}~({\mathsf{ROW}}~0) = x$,
${\mathsf{U}}~({\mathsf{ROW}}~1) = x$
and
${\mathsf{listlen}}~{\mathsf{ROW}} = 2$.
We can then essentially complete the subproof
using the local assumptions
$${\mathsf{P}}~({\mathsf{ap}}~{\mathsf{employs}}~({\mathsf{listset}}~({\mathsf{cons}}~x~({\mathsf{cons}}~x~{\mathsf{nil}}))))$$
and
$${\mathsf{P}}~({\mathsf{SR}}~({\mathsf{listset}}~({\mathsf{cons}}~{\mathsf{employs}}~({\mathsf{cons}}~{\mathsf{uses}}~{\mathsf{nil}}))))$$
along with the translation of the following Merge.kif formula:
{\footnotesize{
\begin{verbatim}
(=> (and (subrelation ?REL1 ?REL2) (instance ?REL1 Predicate)
         (instance ?REL2 Predicate) (?REL1 @ROW))
    (?REL2 @ROW))
\end{verbatim}
}}

To complete the contradiction we prove
$\lnot {\mathsf{P}}~({\mathsf{ap}}~{\mathsf{uses}}~({\mathsf{cons}}~x~({\mathsf{cons}}~x~{\mathsf{nil}})))$.
The three most significant Merge.kif formulas whose translated propositions
are used in the subproof are:
{\footnotesize{
\begin{verbatim}
(instance uses AsymmetricRelation)

(subclass AsymmetricRelation IrreflexiveRelation)

(=> (instance ?REL IrreflexiveRelation)
    (forall (?INST) (not (?REL ?INST ?INST))))
\end{verbatim}
}}
That is, Merge.kif declares that ${\mathsf{uses}}$ is an asymmetric relation,
every asymmetric relation is an irreflexive relation,
and that irreflexive relations have the expected property of irreflexivity.

\section{ATP Problem Set}\label{sec:atp}

% whether we could achieve a correct and consistent semantics for our interpretation.
%We tested whether a number of
%first order problems could be proven in our higher-order translation.  We also generated
%problems to test performance, and also as with \cite{SchulzSutfliffeUrbanPease} to run a 
%variety of synthetically-generated problem to see whether we could find a contradiction,\todo{What synthetically-generated problems do we mean here? AP - I thought that the subgoals are generated automatically from your interactive proof.  Is that not the case?}
%as indicated by a proof that does not use the conjecture.\todo{Adam, I think you wrote this paragraph, so I'm reluctant to just delete it -- but I don't think it fits here. I'm writing like it's deleted and we can decide later where to move it and how to change what it says. It seems redundant in that we already say the problems coming from queries we're using and roughly why. - C. AP- I'm happy for you to delete anything I've written if it doesn't make sense to you :-)}

After interactively proving the 23 problems, we created TH0 problems
restricted to the axioms used in the proof. This removes the need for the
higher-order ATP to do premise selection.
Additionally we used Megalodon to analyze the interactive proof
to create a number of subgoal problems for ATPs
-- ranging from
the full problem (the initial goal to be proven) to
the smallest subgoals (completed by a single tactic).
For example, the interactive proofs of Examples~\ref{ex:wordex}, \ref{ex:tqg22alt4} and \ref{ex:tqg11}
generate 415, 322 and 100 TH0 problems, respectively.
In total analysis of the interactive proofs yields 4880 (premise-minimized) TH0
problems for ATPs.
In Table~\ref{tab:sgresults} we give the results for several
higher-order automated theorem provers (Leo-III~\cite{DBLP:journals/corr/abs-1802-02732}, Vampire~\cite{Vampire}, Lash~\cite{BrownK2022}, Zipperposition~\cite{10.1007/978-3-030-79876-5_24}, E~\cite{Sch02-AICOMM}), given a 60s timeout.
%Given one minute (and told not to use SINE), Lash can
%prove 284 (68\%), 201 (62\%) and 45 (45\%)
%of these subgoal problems, respectively.

%The higher-order automated theorem prover Lash~\cite{BrownK2022}
%can prove 7 of the 23 within 5m (when explicitly told not to use SINE for premise selection)
%including the problems ${\mathsf{TQG3}}$ and ${\mathsf{TQG27}}$ mentioned above.\todo{Which can Vampire and E and Zipperposition do? LEO-II can apparently do 8, but given 10m instead of 5m. Maybe we can just stick with 60s time out and infer how many tops can be done by how many 100 percent on the sgs can be done.}
\begin{table}[htbp!]
  \begin{center}
    {\small{
    \begin{tabular}{lclllll}\toprule
      Problem & Subgoals & Zipperposition & Vampire & E & Lash & Leo-III \\\midrule
${\mathsf{TQG1}}$ & 50 & 50 (100\%) & 50 (100\%) & 50 (100\%) & 50 (100\%) & 50 (100\%) \\
${\mathsf{TQG3}}$ & 20 & 20 (100\%) & 20 (100\%) & 14 (70\%) & 20 (100\%) & 8 (40\%) \\
${\mathsf{TQG7}}$ & 195 & 188 (96\%) & 185 (95\%) & 180 (92\%) & 160 (82\%) & 158 (81\%) \\
${\mathsf{TQG9}}$ & 19 & 19 (100\%) & 19 (100\%) & 19 (100\%) & 19 (100\%) & 19 (100\%) \\
${\mathsf{TQG10}}$ & 112 & 112 (100\%) & 112 (100\%) & 100 (89\%) & 58 (52\%) & 96 (86\%) \\
${\mathsf{TQG11}}$ & 100 & 76 (76\%) & 39 (39\%) & 67 (67\%) & 45 (45\%) & 13 (13\%) \\
${\mathsf{TQG19}}$ & 37 & 34 (92\%) & 22 (59\%) & 20 (54\%) & 37 (100\%) & 11 (30\%) \\
${\mathsf{TQG20}}$ & 41 & 34 (83\%) & 22 (54\%) & 20 (49\%) & 41 (100\%) & 13 (32\%) \\
${\mathsf{TQG21}}$ & 207 & 154 (74\%) & 150 (72\%) & 143 (69\%) & 101 (49\%) & 56 (27\%) \\
${\mathsf{TQG22alt3}}$ & 319 & 246 (77\%) & 214 (67\%) & 193 (61\%) & 197 (62\%) & 136 (43\%) \\
${\mathsf{TQG22alt4}}$ & 322 & 251 (78\%) & 218 (68\%) & 197 (61\%) & 201 (62\%) & 142 (44\%) \\
${\mathsf{TQG22}}$ & 315 & 271 (86\%) & 224 (71\%) & 212 (67\%) & 201 (64\%) & 142 (45\%) \\
${\mathsf{TQG23}}$ & 67 & 61 (91\%) & 67 (100\%) & 42 (63\%) & 51 (76\%) & 38 (57\%) \\
${\mathsf{TQG25alt1}}$ & 910 & 652 (72\%) & 526 (58\%) & 580 (64\%) & 529 (58\%) & 246 (27\%) \\
${\mathsf{TQG27}}$ & 7 & 7 (100\%) & 7 (100\%) & 7 (100\%) & 7 (100\%) & 7 (100\%) \\
${\mathsf{TQG28alt1}}$ & 600 & 428 (71\%) & 386 (64\%) & 349 (58\%) & 261 (44\%) & 213 (36\%) \\
${\mathsf{TQG30}}$ & 4 & 4 (100\%) & 4 (100\%) & 3 (75\%) & 4 (100\%) & 4 (100\%) \\
${\mathsf{TQG33}}$ & 112 & 82 (73\%) & 83 (74\%) & 79 (71\%) & 85 (76\%) & 36 (32\%) \\
${\mathsf{TQG45}}$ & 162 & 258 (159\%) & 131 (81\%) & 128 (79\%) & 106 (65\%) & 36 (22\%) \\
${\mathsf{TQG46}}$ & 344 & 141 (41\%) & 215 (62\%) & 225 (65\%) & 163 (47\%) & 144 (42\%) \\
${\mathsf{TQG47}}$ & 186 & 113 (61\%) & 113 (61\%) & 109 (59\%) & 93 (50\%) & 79 (42\%) \\
${\mathsf{TQG48}}$ & 336 & 249 (74\%) & 234 (70\%) & 219 (65\%) & 184 (55\%) & 146 (43\%) \\
${\mathsf{wordex}}$ & 415 & 315 (76\%) & 255 (61\%) & 236 (57\%) & 284 (68\%) & 143 (34\%) \\\midrule
Total & 4880 & 3765 (77\%) & 3296 (68\%) & 3192 (65\%) & 2897 (59\%) & 1936 (40\%)\\\bottomrule
    \end{tabular}
    }}
  \end{center}
  \caption{Number of Subgoals Proven Automatically in 60 seconds}\label{tab:sgresults}
%  \vspace{-42pt}
\end{table}
%  \vspace{-52pt}

%\todo{AP - are these all 60 second results?}

%We automatically created a number of problems drawn from our interactive proofs.  Automated theorem proving fails to find the solutions that can be constructed automatically.  So, we created problems to pose to automated provers that are subgoals of the interactive proofs.   

%In 600 seconds, LEO-III can solve 4078 out of 4880 problems and in 60 seconds it solves 1936.   In no case did we find a contradiction without using the conjecture.

\section{Future Work}\label{sec:futurework}

The primary plan to extend the translation is to include
temporal and modal operators. SUMO includes many modal
operators including necessity, possibility,
deontological operators (obligation and permission)
and modalities for knowledge, beliefs and desires.
Each modality can be modelled using Kripke style semantics~\cite{Kripke1963}
(possible worlds with an accessibility relation).

The following is an example of a SUMO formula in Merge.kif using modalities:
{\footnotesize{
\begin{verbatim}
(=> (modalAttribute ?FORMULA Necessity)
    (modalAttribute ?FORMULA Possibility))
\end{verbatim}
}}
\noindent
The current translation simply skips these formulas as they are not in the
SUMO-K fragment. If we only wanted to extend the translation to include
necessity and possibility, we could change the translation to
make the dependence on worlds explicit.
The SUMO formula above could translate to the proposition
$$\forall w\in W.\forall \varphi:\iota\to \iota.(\forall v\in W.R~w~v\to {\mathsf{P}}~(\varphi~v))\to (\exists v\in W.R~w~v\land {\mathsf{P}}~(\varphi~v)).$$
Here $W$ is a set of worlds and $R$ is an accessibility relation on $W$.
Note that the translated formula variable has type $\iota\to\iota$ instead of type $\iota$
to make the dependence of the formula on the world explicit.
In general, terms, spines and formulas would depend on a world $w$
and in an asserted formula the world $w$ would be universally quantified (ranging over $W$) as above.

If we took the approach above to model necessity and possibility,
then to add deontic modalities later we would need a second set of worlds
and accessibility relation. The translation of terms would then have type $\iota\to\iota\to\iota$
to account for the dependence on both kinds of worlds.
In order to prevent needing to keep adding new dependencies for every modalities,
our plan is to combine the sets of worlds and accessibility relations
in an extensible way.
Thus terms will translate to have type $\iota\to\iota$ essentially giving
dependence on a single set encoding a sequence of worlds (where we are open ended
about the length of the sequence).
Using this idea, the SUMO formula above would translate to something like
$$
\begin{array}{c}
  \forall w\in (\Pi x\in X.W~x).\forall \varphi:\iota\to \iota.(\forall v\in (\Pi x\in X.W~x).R~m~w~v\to {\mathsf{P}}~(\varphi~v))\\
  \to (\exists v\in (\Pi x\in X.W~x).R~m~w~v\land {\mathsf{P}}~(\varphi~v))
\end{array}
$$
where $X$ is an index set (where each $x\in X$ corresponds to a modality
being interpreted),
$m\in X$ is the specific index for necessity and possibility,
$W~x$ is the set of worlds for $x$, and $R~x$
is a relation between $w,v\in\Pi x\in X.W~x$ that holds if the $x$ components
satisfy the accessibility relation over $W~x$
and the other components of $w$ and $v$ do not change.
This allows us to model an arbitrary number of modalities using
Kripke semantics while only carrying one world argument.
Another advantage is that it minimizes the change to the
translation of formulas in the SUMO-K fragment (without modalities).
The only required change is to add a single dependence on $w$ via
a new argument and universally quantify over $w$ if the formula is asserted.

We have already done some experiments with this approach
and it shows promise. The previous experiments need to be
extended to include changes that have occurred to obtain
the SUMO-K translation described in the present paper.
Once this is done, we must ensure that translated examples
both with modalities and the examples in this paper without modalities
are provable interactively.
We plan to also test automated theorem provers
on the subgoals obtained from the interactive proofs.
Doing so with the 23 examples in this paper will give an indication
how much more difficult the translated problems become if the Kripke
infrastructure to handle modalities is included.

Another aspect of SUMO are modalities involving likelihood and probability.
These cannot be modelled by Kripke semantics (as the modalities are not normal).
We are experimenting with using neighborhood semantics to include these modalities.

\section{Conclusion}\label{sec:concl}

We have described a translation from the SUMO-K fragment of SUMO
into higher-order set theory. We have considered a number
of examples that use aspects of SUMO-K that go beyond traditional first-order logic,
namely variable arity functions and relations,
row variables, term level $\kappa$-binders
and arithmetic.
We have described a number of interactive
proofs of translated queries
and tested higher-order automated theorem provers
on problems obtained by doing premise selection using
the corresponding interactive proofs.
This gives a set of problems for automated theorem provers
that come from the area of ``common sense reasoning,''
an area quite different from the more common sources
of formalized mathematics and program verification.
On most of the examples, higher-order automated theorem provers
cannot fully automatically prove the % premise selected
query,
but they perform reasonably well on subgoal problems extracted
from the interactive proofs. This gives an indication that
the full problems (assuming premise selection)
are not too far out of reach for current state of the art higher-order
automated theorem provers.
%\todo{Chad: OK, I just wrote this a bit quickly. Feel free to completely rewrite it, but at this point it's better to have something in case nothing changes and we just need to delete remaining todo's to submit. AP - looks good to me}

\subsubsection{Acknowledgments}
The results were supported by the Ministry of Education, Youth and Sports within the dedicated program ERC CZ under the project POSTMAN no. LL1902.

\bibliographystyle{splncs04}
\bibliography{sumo2set}
\end{document}